\documentclass[10pt,journal,compsoc]{IEEEtran}
\newtheorem{definition}{Definition}
\ifCLASSOPTIONcompsoc
\usepackage[nocompress]{cite}
\else
\usepackage{cite}
\fi

\usepackage{hyperref} 
\usepackage{graphicx}
\usepackage{amsmath}
\usepackage{amssymb}
\usepackage{color}
\usepackage{epstopdf}
\usepackage{array}
\usepackage{multirow}
\usepackage{amsfonts}
\usepackage{booktabs}
\usepackage{float}
\usepackage{subfigure}
\usepackage{algorithm}
\usepackage{algpseudocode}
\usepackage[switch]{lineno}
\usepackage{comment}
\usepackage{enumitem}
\setlist{leftmargin=5.5mm}

\begin{document}
%
\title{KANDINSKYPatterns - An experimental exploration environment for Pattern Analysis and Machine Intelligence}
%
\author{Andreas Holzinger, Anna Saranti, and Heimo Mueller
\IEEEcompsocitemizethanks{\IEEEcompsocthanksitem This work was supported by the Austrian Science Fund (FWF), Grant P-32554 explainable Artificial Intelligence.
A. Holzinger, A. Saranti, and H. Mueller are with the Medical University Graz, Austria (Email: andreas.holzinger@medunigraz.at, anna.saranti@medunigraz.at, heimo.mueller@medunigraz.at). \protect\\
}
}

\markboth{IEEE Transactions on Pattern Analysis and Machine Intelligence}
{Holzinger, Saranti, and Mueller \MakeLowercase {
}: KANDINSKYPatterns}

\IEEEtitleabstractindextext{
\begin{abstract}

Machine intelligence is very successful at standard recognition tasks when having high-quality training data. There is still a significant gap between machine-level pattern recognition and human-level concept learning. Humans can learn under uncertainty from only a few examples and generalize these concepts to solve new problems. The growing interest in explainable machine intelligence, requires experimental environments and diagnostic tests to analyze weaknesses in existing approaches to drive progress in the field. In this paper, we discuss existing diagnostic tests and test datasets such as CLEVR, CLEVERER, CLOSURE, CURI, Bongard-LOGO, V-PROM, and present our own experimental environment: The KANDINSKYPatterns, named after the Russian artist Wassily Kandinksy, who made theoretical contributions to composability, i.e. that all perceptions consist of geometrically elementary individual components. This was experimentally proven by Hubel and Wiesel in the 1960s and became the basis for machine learning approaches such as the Neocognitron and the even later Deep Learning. While KANDINSKYPatterns have computationally controllable properties on the one hand, bringing ground truth, they are also easily distinguishable by human observers, i.e., controlled patterns can be described by both humans and algorithms, making them another important contribution to international research in machine intelligence.
\end{abstract}

\begin{IEEEkeywords}
Pattern Analysis, Machine Intelligence, dataset
\end{IEEEkeywords}}


\maketitle
\IEEEdisplaynontitleabstractindextext
\IEEEpeerreviewmaketitle


\label{sec:introduction}
\section{Introduction}
\IEEEPARstart{A}rtificial Intelligence (AI) indisputably makes great advances in many tasks from pattern recognition to image analysis -- thanks to the progress in data driven statistical machine intelligence \cite{Jain:2000:statistical}. Recent examples have demonstrated that AI can reach even human-level performance and beyond -- even in complex domains such as medicine \cite{EstevaThrun:2017:DermaNN}, \cite{ArdilaEtAl:2019:DeepSuccess}, \cite{Esteva:2021:DeepVisionNature}. 

However, such models are highly dependent on the quality of the input data and especially on the training data. Although the latter are of eminent importance for learning, they are typically treated as predefined, static information, i.e. the current best models are passive and rely on human-curated training data, but have no control by themselves. This is in contrast to how humans learn, because humans interact with their environment to gain information \cite{MisraVanDerMaaten:2018:LearningAsking}. The role of interactivity, which is especially important for learning new concepts, and the extent to which the learner can take an active role in learning those concepts has been considered extremely important by the machine intelligence research community for a very long time \cite{SammutBanerji:1986:FirstOrderLogic}. Unlike AI, sometimes - of course not always - humans are very good in understanding and explaining concepts, even in novel situations with complex dynamics - even with little interaction \cite{Tenenbaum:2020:Human-Level}. 


Human conceptual abilities are also very \emph{productive}: humans can understand and generate novel concepts through compositions of existing concepts, 
unlike standard machine classifiers which are limited to a fixed set of classes. Moreover humans are able to induce "ad hoc" categories \cite{Barsalou:1983:AdHocCategories}. Thus, unlike AI systems, humans reason seamlessly in large, essentially "unbounded" concept spaces and are very good at dealing with uncertainty and under-determination \cite{Piantadosi2016-bx}. 





This paper is structured as follows. After the introduction and motivation we provide in section \ref{sec:background} a brief overview on some background in concept learning; In section \ref{sec:relatedwork} we give an overview on related work including some benchmark datasets for concept learning, reasoning and generalization. Finally, in section  \ref{sec:KandindskyPatterns} we present our experimental environment called KANDINSKYPatterns (KP) along with three challenges for the international research community. Finally, in the conclusion, we contrast the main properties of KP with current approaches and propose some further future work.

\section{Background}
\label{sec:background}

Wassily KANDINSKY (1866--1944) was an expressionist artist, one of the pioneers of abstract art, and taught at the famous Bauhaus school in Germany between 1922 and 1933, where he promoted simple colors and simple shapes and published in 1926 his book on point and line to surface: contribution to the analysis of painting elements
\cite{Kandinsky:1926:PunktLinie}. In 1959 Hubel \& Wiesel \cite{HubelWiesel:1959:Cat} carried out their famous experiment where they discovered that the visual system of the cat brain builds up an image from very simple elements into more complex representations. Similarly, a deep learning architecture can be viewed as a multilayer stack of simple modules, most of which are subject to learning and many of which compute nonlinear input-output mappings. Each module in the stack transforms its input to increase both the selectivity and invariance of a representation. With multiple such nonlinear layers, a system can implement complicated functions of its inputs that are simultaneously sensitive to detail and insensitive to large irrelevant variations such as background, pose, lighting, and surrounding objects. At first edges and lines are learned, then shapes, then objects are formed, eventually leading to concept representations \cite{LeCunBengioHinton:2015:DeepLearningNature}. 

Humans group segments into objects and use concepts of object permanence and object continuity to explain what has happened and infer what will happen, and also to imagine what would happen in counterfactual situations. The problem of complex visual understanding has long been studied in computer vision \cite{Antol:2015:VQAVersion_1}. However, the underlying factors, especially the causal structure behind comprehension and explanation processes, have been less explored. Therefore, experimental datasets that can be understood by both machines and humans are of great importance to foster future advances of the international research community.

\section{Related Work}
\label{sec:relatedwork}

\subsection{Human versus machine perception in the medical domain}

Several research works consider the similarities and differences of perception capabilities of humans and machine learning models in the medical domain \cite{Makino:2020:DifferencesHumanMachineMedical}. Although both of them can be modelled by the same Probabilistic Graphical Model (PGM) \cite{Koller:2009:ProbabilisticGraphicalModels}, \cite{Saranti:2019:LearningCompetencePGM}, as far as structure goes, they do have significant differences in the learned parameters and predicted outcomes of the posterior variables. Humans are in some cases more robust to perturbations of the input data; it can be shown that machine learning models don't just suffer from dataset shift but also information loss. Furthermore, there is a discrepancy between them, when comparing what parts in the input image were helpful for the diagnosis. By comparison with radiologists diagnoses, researchers can enhance the dataset with images that will occur in medical practice and make the machine learning models more robust to perturbations.

\subsection{Synthetic datasets for benchmarking Concept Learning, Reasoning and Generalization}
\label{DatasetsGeneralization}

One of the first dataset that was used for image classification tasks was the COCO (Common Objects in Context) \cite{Lin:2014:MicrosoftCOCO}, \cite{Chen:2015:MicrosoftCOCOCaptions}. It contained images with objects mostly in their natural surroundings. Machine learning models were challenged with the tasks of object localization as well as prediction of semantic descriptions (captions) of the content. One of the earliest works that learn captions from data does not use neural networks but Conditional Random Fields (CRF) \cite{Kulkarni:2013:Babytalk}. In the case where neural networks are used, the general architecture to solve this task contains a Convolutional Neural Network (CNN) that processes the image and extracts feature embeddings (usually corresponding to particular regions) that will be used as input to a bidirectional Long Short-Term Memory (bi-LSTM) network that aligns those visual embeddings with the embeddings of the caption \cite{Karpathy:2015:DeepVisualSemanticAlignments} \cite{Hendricks:2016:GeneratingVisualExplanations}. The prediction of the semantic description is based on the correlation of the image and with a set of possible captions, which are considered weak labels. Further research demonstrated the role of the image context (surroundings) in the performance of those methods \cite{Lai:2019:Contextual}, specially for the objects are relatively small compared to the others as well as partially occluded by others. Explainable AI (xAI) methods, such as Layer-wise Relevance Propagation (LRP), also showed how artefacts in datasets can be misused by those models \cite{Lapuschkin:2019:CleverHans}. 




After the captioning tasks reached a desirable performance with neural networks, one research direction proceeded into Question Answering (QA) systems, where corresponding datasets such as the Visual Genome \cite{Krishna:2017:VisualGenome}, also evolved. Instead of hard or soft labelling, it was relevant to find out if neural networks are capable of answering user posed questions about some property of objects in the image. For an Artificial Intelligence (AI) system to support human dialogue is more natural and more informative, since some information is encoded in the question \cite{Mao:2019:NeuroSymbolicConceptLearner}. But for a neural network being able to answer correctly a set of questions, following some pre-specified structure and usually increasing complexity, means that it can handle concept learning as well as having reasoning abilities, which go beyond embeddings alignment. In general, questions and answers in future research will depart from template-generated constrained versions, will be longer, contain more combinations of objects, concepts and ideally will be free-text \cite{Agrawal:2017:CVQAACompositionalSplit}. Same applies to images that depict more real-life elements than carefully constructed rendered scenes as well as videos. Currently, video question answering systems (VideoQA) are evolving, with the laborious task of gathering appropriate data \cite{Yang:2020:JustAskVideoQA}. An overview of VQA systems, their properties and abilities is provided by \cite{Kojima:2020:WhatisLearnedVQA}.

The generalization ability of visual question answering and concept learning systems is also a characteristic that needs to be quantified by well-designed splits of the dataset \cite{Agrawal:2017:CVQAACompositionalSplit}. Adding noise as well as variability in the images is a first step towards improving cross-dataset generalization, even if the learned models don't have increased performance \cite{Torralba:2011:DatasetBias}. Appearance variability is ensured by gathering data from independent resources \cite{Lin:2014:MicrosoftCOCO}, \cite{Karpathy:2015:DeepVisualSemanticAlignments} and captions must be reviewed by several people \cite{Lin:2014:MicrosoftCOCO}, \cite{Chen:2015:MicrosoftCOCOCaptions}.  Ideally, the invented model architectures learn disentangled representations of the concepts at training time and then they are able to compose those concepts at test time, even if those are extremely rare or even physically impossible in real-world settings. Furthermore, they should be able to interpolate, extrapolate and obtain some abilities of zero-shot learning as for example being able to count more objects than the ones encountered in all images of the training set or to recognize a never seen before colour. A neural network is considered to be even more capable if the training set contains only a few of the possible combinations \cite{Bahdanau:2019:SystematicGeneralizationWhatIsRequired}.  

The SAR Altimetry Coastal \& Open Ocean Performance (SCOOP) dataset \cite{Bahdanau:2019:SystematicGeneralizationWhatIsRequired} is one that is not tailored particularly in concept learning, but concentrates on exercising the generalization capabilities of Neural Modular Networks (NMN) \cite{Andreas:2016:NeuralModuleNetworks} \ref{NeuralModuleNetworks} with different layouts that do not need prior knowledge about the task, and Memory, Attention, and Composition (MAC) \cite{HudsonManning:2018:MACCompositionalAttention}, Relation Networks (RN) \cite{Santoro:2017:SimpleNNRelationalReasoning} \ref{RelationNetworks} and Feature-wise Linear Modulation (FiLM) \cite{Perez:2018:FiLM}. The proposed dataset contains images with letters and digits and the reasoning tasks encompass only spatial relations. The experiments showed that the architecture of NMNs played a substantial role in their generalization capabilities; modules that are organized and connected in tree structure have an excellent generalization performance, comparable with models that use prior knowledge, whereas a set of modules structured as a chain fail just because of that reason. It is a key insight that to achieve compositionality, apart from parameterization their has to be successful specialization between the modules and the layout must be inducted adequately.

Recent research of Keysers et al. \cite{Keysers:2020:MeasuringCompositionalGeneralization} provides directives on how to systematically construct splits of synthetic datasets that have a primary goal to measure compositional generalization and not some domain adaptation. According to their research, each different training - test split consists of a different compositionality experiment; they exercised different encoder-decoder architectures and showed that cases with overall very high accuracy have a significant drop in performance (reaching even values as low as $20\%$ accuracy) for carefully designed splits. They adapted an already created textual dataset comprised by very few atoms and rules, thereby enforcing that complexity will only emerge as a result of rule composition. Desired properties of benchmark datasets should be that atoms and their distribution are similar in both training and test set, but the distribution of the compositions is at the same time as different as possible (as measured by the Chernoff coefficient \cite{Chung:1989:DistanceProbabilityDistributions}). The designed dataset, as well as the splits, do exercise generalization abilities of previously invented ones, such as extrapolation, number of patterns and so on. Nevertheless, the images in each split are not generated fulfilling only one criterion, but several ones at the same time.

\subsubsection{Visual Question Answering dataset (VQA)}
\label{VQA}

One of the first visual question answering datasets is simply called Visual Question Answering (VQA) and was created in 2015 \cite{Antol:2015:VQAVersion_1}. The dataset consisted of real images as well as questions that were created based on human-generated captions. State of the art architectures of that time provided good performance, but their generalization and compositionality was questioned and tested with an extension of the original dataset called Compositional VQA (C\_VQA) \cite{Agrawal:2017:CVQAACompositionalSplit}. This dataset was carefully crafted so that the distribution of questions across splits (in comparison to its first version \cite{Antol:2015:VQAVersion_1}) remains the same across splits, but the answer distributions for a particular question type should be different. All architectures showed a drop in performance on the new dataset - even the Neural Module Networks \cite{Andreas:2016:NeuralModuleNetworks}, \ref{NeuralModuleNetworks} which have a built-in compositional architecture. This is assumed to be because of the LSTM that uses strong language priors, which in the case of the original dataset are similar for training and test set, but in C\_VQA this was not longer the case.

\subsubsection{Compositional Language and Elementary Visual Reasoning diagnostics dataset (CLEVR)}
\label{CLEVR}

The Compositional Language and Elementary Visual Reasoning diagnostics dataset (CLEVR) \cite{Johnson:2017:Clevr} is one of the mostly used datasets that were used for visual question answering systems and was also challenged and improved by diverse research groups \cite{Kim:2018:NotSoCLEVRL}. The scene contains preliminary simple three-dimensional geometrical objects (cubes, cylinders, spheres) of various colours, sizes and textures, each time in a different constellation. For each image there is a set of possible questions that were synthetically generated with a functional program; each of them is created by a chain or tree of reasoning functions encompassing mainly relations, logical, existence, uniqueness, counting and comparison concepts. By that means, the ground truth is available on-demand and can be used to verify the performance of neural networks that learn the corresponding concepts which are presupposed for resolving the reasoning tasks and correct answering of questions. Surprisingly, there were technical solutions \ref{NeuralModuleNetworks} that provided the expected answer, without necessarily following the ground-truth defined reasoning sequence. 

The generalization aspect \ref{DatasetsGeneralization} was considered thoroughly in CLEVR; rejection sampling was incorporated to make sure that the answer distribution is uniform. Furthermore, the questions textual content can impose biases, since long and complex questions contain more concepts and will need longer reasoning paths on-trend. The neural network architectures of that time relied on image and textual alignment of embeddings and therefore showed poor generalization results. Even in cases where the test image was only differing by one attribute value from an image that was encountered in the training set, the performance was not satisfactory. This indicated the fact that deep learning models do not have disentangled representations of attributes and objects which explained the drop in performance of those models on unseen images and questions that were composed by the same means (distributions) as the ones in the training set. 

An expansion of CLEVR called CLOSURE was invented in 2020 by Bahdanau et. al. \cite{Bahdanau:2019:Closure}, together with a proposed architecture that is inspired by Neural Module Networks \cite{Johnson:2017:Clevr} \ref{NeuralModuleNetworks} as well as symbolic approaches \ref{Neurosymbolic}, described in \ref{Neurosymbolic}. They observed that CLEVR does not have uniformity in the label of the training and test set, comparison questions only use spatial referring expressions and other related artefacts that made the path for the creation of an enhanced synthetic dataset. They argue that the questions in the test set must have a different distribution - even if the image distribution is the same - and it should contain new combinations of already learned semantic and syntactic components. Another variation of the CLEVR dataset concentrating on relational reasoning is Sort-of-CLEVR: \cite{Santoro:2017:SimpleNNRelationalReasoning}. This is a distilled two-dimensional version of CLEVR containing images that have always a fixed number of objects, the only attributes are colours, the questions have fixed length and consist of relational and non-relational questions. Another variant called CLEVR-Humans \cite{Johnson:2017:ClevrHumans} contains the same images but uses questions that are posed by humans, have much more diversity and are linguistically more complex than the synthetic generated ones.

\subsubsection{CoLlision Events for Video REpresentation and Reasoning dataset (CLEVERER)}
\label{CLEVERER}

CoLlision Events for Video REpresentation and Reasoning (CLEVERER) \cite{Yi:2019:Clevrer} is a dataset that extends CLEVR \ref{CLEVR} with the goal of exercising temporal and causal reasoning abilities of neural networks. The dataset is not comprised of static images, but of videos that depict collisions between the predefined objects, which as in CLEVR have a predefined set of attributes and appear in different constellations. To answer the questions provided in the dataset correctly, the neural networks have to be able to reason counterfactually, in terms of ``what-if'' and ``what-if not'' a collision event would happen (or not) and explaining which dependencies between positions and occurring events exist. Recognizing the motion of the moving objects in the videos helps the models maintaining non-static but constant information about the involving objects. Causal relation recognition requires separate object representations, whereas causal reasoning could be overtaken by symbolic logic \ref{Neurosymbolic}, resembling roughly the System 1 and System 2 division of cognitive abilities \cite{TverskyKahnemann:1974:JudgementUncertainty}, \cite{Kahneman:2001:thinkingFastSlow}. As in CLEVR, counteracting bias was supported by ensuring that each possible answer of each question is valid in the same number of images.

\subsubsection{Compositional Reasoning Under Uncertainty dataset (CURI)}
\label{CURI}

Newer benchmark datasets are created with the goal of exercising machine learning models on concept learning under uncertainty \cite{Vedantam:2020:CuriConceptLearningUncertainty}. Concepts can be ambiguous and are no longer rigid labels; for example, one image can belong to many different concepts. Each concept is expressed by a probabilistic context-free grammar containing (among others) logical operators and comparisons. The number of samples in the dataset is predefined so that the acquisition of a concept has to be made by limited data. Furthermore, new challenges arise by the comparison of different data modalities, such as images, sounds and symbolic schemes in textual form. The comparison of the different performances, representations and appropriate models sheds light on the commonalities and differences of the input data and the task itself.

The data are according to pre-defined hypotheses and either satisfy a concept (positives) or not (negatives), with a particular probability. Current representation learning methods  \cite{Kipf:2019:ContrastiveLearning} that are used to embed the states of reinforcement learning environments, are also based on this scheme of positive and negative examples created by corruption of the positives. The compositionality gap is measured by the comparison between an ideal Bayesian learner that has access to all the hypotheses. Furthermore, stronger and weaker generalization tasks are defined, by the targeted choice of negatives that have partial overlaps with the concept of the positives. Generalization is tested by strategically designed splits between the data that are used for training and ones that comprise the test set. For example, to evaluate how well the generalization is accomplished, easier concepts (with smaller prefix sequence length) are present only in the training set, whereas the test set 
contains only complex concepts.

\subsubsection{Raven-Style Progressive Matrices (RPM) dataset}
\label{RPM}

The Raven Progressive Matrices (RPM) test \cite{Carpenter:1990:RavenProgressiveMatricesTest}, \cite{Raven:2000:Rpms} is a non-verbal test, invented by psychologists mainly to test the recognition of relations between objects and attributes. Each test is comprised of a $3 \times 3$ matrix; each row of this matrix contains images having the same relationship with each other. The relations consist of logical operators, comparisons and counting objects, which is still a challenging task for current state-of-the-art neural networks. The questioned entity must choose the adequate answer from a set of candidate images; this consists a major difference with the rest of the datasets and cognitive tasks explained in this section. The images consist of simple abstract shapes, selected from a closed set. A smaller but more diverse dataset following the main principles of RPM, called Relational and Analogical Visual rEasoNing (RAVEN) \cite{Zhang:2019:RavenDataset}, was used to exercise the reasoning capabilities of Relation Networks \ref{RelationNetworks}. The creation of each image is made with the use of more structures and instantiations, as well as follows more diverse types of rules - although they are hierarchical. Furthermore, researchers did not only compare the performance of the models w.r.t. ground truth, but also to human performance.

A new benchmark for visual reasoning for real images goes beyond testing the generalization capabilities of non-abstract scenes \cite{Teney:2020:VPROM}. Although the style of the posed problem follows the structure of the RPMs, meaning that the input is still a $3 \times 3$ matrix of images connected by a particular relation at each row,  the set of possibilities is open and the data are sampled from the Visual Genome \url{https://visualgenome.org/}. As for the purpose of CLEVERER \ref{CLEVERER}, the ultimate goal is to evolve towards real data scenarios that will be of more practical use to the corresponding scientific communities and industry. The researchers of \cite{Teney:2020:VPROM} created a detailed list of data splits with interpolation/extrapolation, held-out objects, attributes and relationships, since the neural networks that they used to struggle with out-of-distribution data. An important requirement for the machine learning solutions that will solve those tasks efficiently is also the computation of a simple abstract description that will be used as a generative explanation for the discovery of the correct image. 

\subsubsection{Bongard-LOGO dataset}
\label{Bongard-LOGO}

The Bongard-LOGO diagnostic dataset \cite{Nie:2020:BongardLogo} has as starting point the Bongard problems (BPs) \cite{Bongard:1968:TheRecognitionProblem}, \cite{Bongard:1970:PatternRecognition}. The Bongard problems are comprised of a set of visual concept learning tasks, each of them defined by two sets of image examples that need to be differentiated. The first one is called positive and the second one negative. All images of the second set do not obey the concept of the first. A textual description of the concept that generated the image sets is desirable; nevertheless, the authors state that there are concepts that are not easy to be expressed even by humans. Therefore it is not a central point of this dataset, since a human or an algorithm could just state if an unseen image belongs to the same concept that generated a fixed number of others, even without explicitly stating what concept that is.

The two sets of image examples have a small number of images created by programs written in the action-oriented LOGO language \cite{Harvey:1997:Logo}. There are fundamental differences between these datasets and the previously mentioned in subsections \ref{CLEVR}, \ref{CLEVERER}, \ref{CURI}. Firstly, although the generated images are comprised of basic elements, such as strokes, have different attributes and belong to different categories, they should not be differentiated by them. There is only one object in the image, but the perception must be rotation-invariant and scale-invariant. Secondly, the fact that different images can be generated by different concepts is not perceived as ambiguity as in CURI \ref{CURI} but as a contextual hierarchical difference. The context is defined by the rest of positive and negative images in the dataset, and one of the main hypotheses is that current pattern recognition machine learning models have a context-free implementation policy.     

One notable similarity of the Bongard-LOGO problems and the third challenge of the KANDINSKYPatterns \ref{challenge_nr_3}. A set of (typically smaller) objects are not perceived individually, but as a whole, provided that they have a recognizable arrangement. (TODO: Reference to image). In this case, there is a trade-off of concepts that are made by analogy making; the image description is not made by the listing of those objects, but the description of the form that is created by all of them. 

The researchers did experiments with current state-of-the-art machine learning models that tackle few-shot problems, use meta-learning principles and symbolic methods to solve the aforementioned tasks. Furthermore, they conducted studies with humans of two levels of expertise to compare the performance of classification. Thereby, they showed that even the most progressed machine learning solutions did not perform as well as humans; this is an indication of the lack of concept learning capabilities compared to humans. For this type of research it is important not to find a solution to a particular subset of problems - since some Bongard problems are already solved - but an effective solution that will encompass all posed problems.

\subsubsection{Other related diagnostic datasets}

A two-dimensional dataset that has several similarities with the KANDINSKYPatterns is "ShapeWorld" \cite{Kuhnle:2017:Shapeworld}. The images do contain abstract shapes and the captions are synthetically generated by following the rules of a predefined grammar. This grammar defines a one-to-one mapping of entities to nouns, attributes to adjectives and relations. The researchers did not proceed in evaluating the performance of state-of-the-art neural networks and comparison with human performance. Furthermore, the dataset "bAbi" for text comprehension \cite{Weston:2015:bAbi} also follows several principles as the aforementioned ones, but targets only textual data. Parallel research for the generation of synthetic datasets for physical \cite{Bakhtin2019-sy} and mathematical reasoning \cite{Saxton:2019:MathematicalReasoning} is currently evolving.

\subsection{Technical Solutions}
\label{TechnicalSolutions}

This section lists and describes the main directions of research that are used currently in concept and representation learning. The list is by no means exhaustive, but it is representative as far as categorization is concerned; variations and improvements of (what is considered to be) the central idea, architecture and overall solution strategy can be   Although distinct, they are sometimes entwined since in several cases components from one technical solution are used in another, usually with an adequate adaptation and improvement. 

\subsubsection{Neural Module Networks (NMN)}
\label{NeuralModuleNetworks}

Neural Module Networks (NMN) \cite{Andreas:2016:NeuralModuleNetworks}, \cite{Andreas:2016:ComposeNeuralNetworksQA}, \cite{Hu:2017:LearningToReason} were the first attempt to solve visual question answering tasks in a modular way. The researchers understood that the questions can be decomposed into concepts that reoccur and thought of specially designed neural network architectures that specialize to each of them separately. The modules themselves contain fully connected, convolution and attention components as well as non linear activations that are composed in a different sequence, depending on the sub-task they need to solve. For example, to combine two already learned concepts, the corresponding module merges the two attentions from the already computed visual groundings by stacking first, then applying convolution and lastly passing the result from a non-linearity. The possible inputs and outputs of the modules are also constrained to be either images, attentions or labels, in a way that resembles software Application Programming Interfaces (API) \cite{Kim:2018:ProgressieModuleNetworks}.

To answer a question, the right modules must be picked and a combined, modular architecture needs to be created. At training time, a separate neural network learns how to select the set of necessary modules, how to connect them with each other so that they jointly learn the necessary representations to answer the question. This can be a recurrent neural network (RNN) that outputs the textual symbolic expression of the optimal module structure. The search over the space of all possible layouts, was made even more efficient with the use of reinforcement learning \cite{Sutton:2018:ReinforcementLearning} and incorporation of expert policies that were used for pre-training.

The neural modules did not learn disentangled representations of concepts, needed one dedicated module for each concept and at test time did not show robustness to unseen questions. Nevertheless, they are considered interpretable, have modular structure meaning that they can be ``pluggable'' in several architectures and were an important starting point for later architectures and research directions such as the Relation Networks (RN) \ref{RelationNetworks} and the neuro-symbolic hybrid models \ref{Neurosymbolic}. The communication between modules in an interface-like way, by the same means, that programs for the composition of higher-order concepts from simpler ones is a recent improvement that makes them more extendable \cite{Kim:2018:ProgressieModuleNetworks}.

\subsubsection{Hybrid Neurosymbolic}
\label{Neurosymbolic}

Neuro-symbolic methods divide their architecture into two parts; one is processing the data learning the appropriate embeddings, whereas the other learn symbolic representations of the input data. The representation of the image is thereby disentangled from the symbolic execution engine that processes its symbolic representation, thereby enabling different types of images to be executed by it as long as they can conform to the learned representation scheme. 

One of the first works \cite{Yi:2018:NeuralSymbolicVQA} parses the scene with a state-of-the-art image processing CNN, to extract objects as well as their features (color, shape, size, material and coordinates). The input question is used as input to an LSTM that outputs programming code to process the structural representation extracted from the image by the CNN. Each logical operation and relation ability required by the model has its corresponding programming language module; the set of all those modules is pre-defined by human domain knowledge and is therefore also interpretable. To test generalization, researchers applied the learned model on a dataset containing Minecraft \url{https://www.minecraft.net/en-us/} scenes where the resulting structural representations were, as expected, longer than the ones that were generated for the CLEVR dataset \ref{CLEVR}. Generalization to images containing natural scenes that were not encountered at training time is not possible with this method.

To make the representation trainable and less rigid, one has to make it learnable by optimization  \cite{Mao:2019:NeuroSymbolicConceptLearner}. The main difference is that after the image is processed by the CNN, neural operators implemented by simple linear layers of neurons learn the concept embeddings. With the use of curriculum learning the easiest concepts involving shape and colour are learned and separated first, and then the neural operators can learn more difficult ones. The optimization is guided by the REINFORCE algorithm \cite{Williams:1992:REINFORCE} to produce a program that obeys a specially designed domain language (DSL) and is therefore interpretable. This helped the overall architecture to have greater generalization capabilities that are tested on scenes containing more objects, non-encountered attribute combinations as well as learning a completely new colour (zero-shot learning). An extended version \cite{Han:2019:VisualConceptMetaconcept} that performs classification by discriminating if two concepts have a particular meta-concept relation, which although needed an enhanced input dataset considering meta-concepts, produced disentangled representations of concepts and was more data-efficient.

In \cite{Bahdanau:2019:Closure} it is shown that for the extension dataset CLOSURE of CLEVR \ref{CLEVR}, neuro-symbolic solutions do not generalize well. The user's domain knowledge is a requirement for the design of this system and for any new dataset the adaptation overhead is bigger and non-systematic compared to other technical solutions.

Neural grammar induction with the use of visual groundings is a state-of-the-art research topic where improvements are constantly discovered \cite{Zhao:2020:VisuallyGroundedCompoundPCFGs}. With the use of a mixture of probabilistic context-free grammars called compound (C-PCFGs), and additional loss for unlabeled text, several performance metrics are improved and biases from which shorter textual descriptions were benefited are weakened. Each production rule of the grammar is followed with a particular probability, which is assumed to have learnable parameterization. Neuro-symbolic methods help beyond supervised visual question answering; they can also provide solutions for automatic syntax and semantics learning from captions, in an unsupervised way \cite{Shi:2019:VisuallyGroundedNeuralSyntaxAcquisition}. Visual grounding of images and sampled constituency trees representations is learned in alternation with a loss that encourages their alignment. The REINFORCE algorithm \cite{Williams:1992:REINFORCE} is used as in \cite{Mao:2019:NeuroSymbolicConceptLearner} for gradient estimation of the parameters; the reward uses the concreteness of the representation of each text constituent and discourages abstract ones that do not have the corresponding grounding. Furthermore, data from other modalities, such as different languages do help the performance of the algorithm.




\subsubsection{Relation Networks (RN)}
\label{RelationNetworks}

Relation Networks \cite{Barrett2018-xv} consist of a specially designed architecture that has high performance on the Sort-of-CLEVR dataset \ref{CLEVR} that focuses on exercising specifically the relational reasoning capabilities. The parameters of the neural network learn relations between the objects in a way that draws parallels with the way CNNs learn weights to capture the translaltion invariances in the input images or the dependencies between the input sequence in the case of Recurrent Neural Networks (RNN). Each pair of objects is linearly weighted, consisting of the input to an individual non-linear function; the sum of all of them is, in turn, parameterizable, making the model considerably complex (quadratically) w.r.t. the number of the number of objects in the image. The authors draw parallels on their solution with graph theory since their model operates on a complete graph; in later solutions \ref{GNNs} of concept learning, RNs are also seen as Graph Neural Networks (GNN) \cite{Saqur:2020:MultimodalGNNs}. The model uses an image processing part and the CNN feature maps are used as input to the relational network; at the same time, the question is processed by an LSTM that conditions the RN with question embeddings; the relations are learned independently from object recognition. The results indicate that the successful architectures for the solution of those tasks must contain separated components for input structure processing and dedicated modules for relational reasoning.

The performance and robustness of Relation Networks (RN) was improved recently by ``pluggable'' modules called Set Refiner Networks (SRN). The main idea bases on the acknowledgement that the effectiveness of a neural network depends on the vector representations of the input elements compute at the perceptual stage (which in the case of images is usually learned by a CNN). SRN modules consist of a stage between the input embedding and the reasoning component, but instead of mapping the embedding to a set, they encode the set representation to an input embedding, that they can be pass onto the RN. Those representations are shown to be decomposed properly and support thereby the relationship learning task, even in cases where an input entity belongs to many set elements. Iterative inference refines an initial output of a set generator to search for its mapping to an appropriate embedding in an unsupervised fashion. Experiments in the image processing domain, comparing the number of derived set elements and the number of objects, as well as tasks considering translations measure the effectiveness of this representation. The Sort-of-CLEVR \cite{Santoro:2017:SimpleNNRelationalReasoning} extension of CLEVR  \ref{CLEVR} is shown to be solvable with higher performance with the use of the SRN module, without any other quantitative changes in the RN architecture. Furthermore, SRNs have shown their value in Reinforcement Learning representation learning of the environment's state and in textual relations detection where the set encoder uses a Graph Neural Network (GNN) to create an iteratively refinable graph vector representation.  

\subsubsection{Graph Neural Networks (GNN)}
\label{GNNs}

Graphs and Graph Neural Networks (GNN) are currently used extensively in image segmentation \cite{Zhou:2019:CgcNet}, text processing \cite{Schnake:2020:xaiGNNs}, biological network analysis \cite{Muzio:2020:BioNetworkAnalysisDL} and xAI methods \cite{Kori:2020:DNNstoConceptGraphs}. More specifically, concept graphs \cite{Kori:2020:DNNstoConceptGraphs} consist of an attempt to compute a graph from the concepts that are learned by trained neural network models and relations thereof. They draw inspiration from Bayesian models (which are also graphical models \cite{Koller:2009:ProbabilisticGraphicalModels}, \cite{SarantiEtAl:2019:graphicalModels}) that have interpretable random variables but lack performance and abilities to generalize to group the weights of trained deep neural networks with hierarchical clustering. The ultimate goal is to find active inference trails in the created graphical model, based on assumptions about the network weights and create visual trail descriptions that will be validated by medical professionals \cite{Holzinger:2016:InteractiveMLHumanInTheLoop}. 

Compositional generalization in both CLEVR and CLOSURE datasets \ref{CLEVR} has also been achieved with high performance using multimodal GNNs \cite{Saqur:2020:MultimodalGNNs}. The caption or question text, as well as the image scene, consist of the two modalities that are represented as graphs; the common GNN (which is a Graph Isomorphism Network (GIN)) calculates the correspondence between them. Joint learning of the representations by fusion is beneficial over symbolic approaches discussed in \ref{Neurosymbolic} for scaling to more natural images, with more complex objects and longer text as well as joint compositional reasoning. The learned GNN embedding is used for downstream tasks, such as caption truth prediction tasks and generalization tests, not only with good overall performance but specific to all different concept learning subtasks. 

An effective solution for the RPM matrices \ref{RPM}, as well as the Euler Diagram Syllogism \cite{Sato:2015:EulerDiagrams}, is given by \cite{Wang:2020:AbstractDiagrammaticReasoningMutliplexGN}. The performance of this solution is better than the previously developed Relation Networks (RNs) \cite{Barrett2018-xv} (described in \ref{RelationNetworks}), although the ultimate goal of this work is also to capture the relations between the objects of different images. The loss that is minimized is identical to \cite{Barrett2018-xv}, but the methodology uses multiplex graph networks that process relations embeddings. One of the key ideas is that the graph does not have as nodes the objects like a scene graph, but the summarization of graphs. The overall architecture contains a pipeline of object representation, graph processing and reasoning; each of them passes the embeddings to the next one. To enable the algorithm to succeed in several different concept learning tasks, different aggregation types are used; concepts that compare size use maximum and minimum feature aggregations, whereas sum is going to be necessary for the counting of objects. It is argued that symbolic methods as the ones described in subsection \ref{Neurosymbolic} would not be effective in the PGM and RAVEN dataset \ref{RPM}, since those datasets do not provide a question to trigger the creation of a grammar or program. Nevertheless, the need for logical rule extraction from the learned entangled representation is explicitly stated as a requirement for interpretability. 

Overall, GNNs are a promising direction for further research for concept and representation learning as well as visual question answering, since they provide new desirable properties that previous neural network architectures did not have. For example, neuro-symbolic methods \ref{Neurosymbolic} can be improved with the application of GNNs \cite{Yang:2020:ObjCentricVisualReasoning}, \cite{Lamb:2020:GNNsSymbolicAI} by processing the probabilistic scene graph extracted from the image, in a different way than the causal models described in \ref{CausalModels}. Furthermore, multi-modal GNNs can express counterfactuals \cite{Holzinger:2021:MultiModalCausabilityGNN} that have been shown profitable for visual question answering solutions \cite{Chen:2020:CounterfactualVQA}, enhancing particularly the generalization capabilities of the models \cite{Gokhale:2020:MutantVQA}.

\subsubsection{Causal models}
\label{CausalModels}

Causal generative models are also used to infer the scene generation process of benchmark datasets per se \cite{VonKluegelgen:2020:CausalGenerativeSceneModels}. Prior knowledge in the form of assumptions and inductive biases about their dependencies and form, are encoded by the pre-defined structure of a probabilistic graphical model (PGM) that uses variational inference as a means to learn its parameters from data. After this unsupervised model is fixed, a competition between the mixture elements is performed and the most likely composition of objects emerges as the result. The use of attention \cite{Burgess:2019:Monet} helps selecting the image regions containing objects and can deal with occlusion as well as scene depth. Thereby, all possible compositions of objects as well as their attributes and relations are sequentially ``explained away'' and all recombinations of their representations are supported in the generative phase. The use of such models for explainability - which is considered built-in per design - and concept as well as representation learning is a promising research direction. 

The work of Hudson and Manning \cite{HudsonManning:2018:MACCompositionalAttention} and \cite{HudsonManning:2019l:NeuralStateMachineProbabilisticGraph} over the years is also concentrated on solving the concept learning and reasoning challenges with the use of causal models. Their research focuses on the computation of a Neural State Machine that extracts a probabilistic scene graph from each input image and expresses thereby the objects, attributes and relations. The probabilistic model will be able to answer questions by applying inference on the causal model in a sequential fashion. Each question is decomposed to reasoning parts; an inference procedure needs to be applied by traversing the causal variables involved to provide the answer. This model also uses domain knowledge, since the semantic concepts are pre-defined and can be used for the factorization of the model. This is exactly what provides the desired disentanglement properties and the required modularity. The embeddings of objects, attributes and relations are defined in an initial state and the goal of the training procedure is to align the degree of belief for each detected component of the image with the corresponding embedding. Since each entity in the image is represented by a set of vectors, the created representations are disentangled and can be recombined at test time on datasets with different distributions, context and text conforming to different grammatical rules with good generalization properties. On the other hand, rarely encountered entities cannot have a good representation, thereby hindering generalization performance. The random variables do provide greater interpretability, since they express known components, but do not support completely unseen configurations that do not have a corresponding random variable properly.

\subsubsection{Reinforcement Learning (RL) inspired solutions}

The work of Misra et al. \cite{Misra:2018:LearningByAskingQuestions} is inspired by design principles of Reinforcement Learning (RL) \cite{Sutton:2018:ReinforcementLearning} to tackle the concept learning problem of CLEVR dataset \ref{CLEVR} not by means of a specially designed neural network architecture, but by learning to adjust the dataset presented to the model during training. The architecture consists of a typical image processing neural network and a text processing one that is conditioned on the extracted image features of the first. Nevertheless, the training procedure does not rely on a ``passive'' dataset; the model even learns the language model that the questions should follow by a process is called Visual Question Generation (VQG). The overall architecture consists of a question generator that is trained to compute questions that are relevant so that the answering module can learn a policy that is driven by rewarding positively the expected accuracy improvement; this expresses the informativeness value of the question that was selected.

The researchers showed that the concepts occurring during the learning phase proceed from the easiest ones to the most difficult ones, which resembles curriculum learning which was explicitly used in neuro-symbolic solutions \ref{Neurosymbolic} and in this work it is emerging. Furthermore, this method is more sample efficient, learns autonomously which questions are more profitable to ask long-term and has an actionable behaviour on discovering which questions are invalid, redundant or more difficult than appropriate. The results showed the model although trained on a dataset that does not have the same distribution in the training and validation set as CLEVR does, it has comparable performance with explicitly designed models trained on CLEVR. In the case of CLEVR-Humans \ref{CLEVR} where the distribution is not the same, it has better performance, thereby indicating increased generalization capabilities \ref{DatasetsGeneralization}.

\subsubsection{Cognitive xAI}
\label{CognitiveXAI}

Cognitive xAI deals with the generation of cognitive rule-based explanations \cite{Rothman:2020:HandsOnXAI}. It can be combined with already established xAI methods and enhancing them by bringing the human-in-the-loop \cite{Holzinger:2016:InteractiveMLHumanInTheLoop}. Domain experts define their own cognitive dictionary with a content that is independent of the xAI method. Nevertheless, the dictionary must be tailored to the method that will be improved and cannot be yet model-agnostic. In this case, the machine learning model is already built with the human-defined rules already, thereby enabling explanations understandable by humans per design.

\section{KANDINSKYPatterns}
\label{sec:KandindskyPatterns}

KANDINSKYPatterns consist of KANDINSKYFigures that are mathematically describable, simple, self-contained, and thus controllable test datasets. Importantly, while they possess these computationally manageable properties, they are also easily distinguishable by human observers. 
Consequently, controlled patterns can be described in experiments by both humans and computers, e.g., as a quasi "intelligence test" for machines \cite{Holzinger:2019:Kandinsky}. We define a KANDINSKYPattern as a set of KANDINSKYFigures, where for each figure an "infallible authority" defines that the figure belongs to the Kandinsky pattern. With this simple principle, we generate training and validation datasets that can be used for a wide variety of purposes, such as testing the quality of explanations and concept learning. Simplicity is necessary to limit the size of the training set. 

\subsection{Definitions}
\label{definitions}

\begin{definition}
A \textbf{KANDINSKYFigure} is a square image containing \textit{1} to \textit{n} geometric objects; \textit{n} limited to either the computational capability and/or limited to a number still distinguishable for the human observer. Each object is characterized by its shape, color, size and position within this square. Objects do not overlap and are not cropped at the border. All objects must be easily recognizable and clearly differentiable by a human observer. 
\end{definition}

The set of all possible KANDINSKYFigures 
is given by definition 1 with a certain set of values for shape, color, size, position and the number of geometric objects. For example we can use for shape the values circle, square, triangle, etc., and for color values such as red, blue, yellow, etc., and we allow arbitrary positions and size with the restriction that it is still human recognizable. Refer to our Website and GitHub repo \url{https://github.com/human-centered-ai-lab/app-kandinsky-pattern-generator} for examples and for some example images to \cite{MuellerHolzinger:2019:arXivKANDINSKY}.

\begin{definition} A Statement \textbf{$s(k)$} about a KANDINSKYFigure $k$ is either a mathematical function, $s(k) \to B$; with $B (0,1)$ or a natural language statement, which is either true or false. 
\end{definition}

Note: The evaluation of a natural language statement is always done in a specific \textit{context}. In the followings examples we use well known concepts from human perception and linguistic theory. If {$s(k)$} is given as an algorithm, it is essential that the function is a pure function, which is a computational analogue of a mathematical function. 

\begin{definition}
A KANDINSKYPattern is defined as the subset of all possible Kandinsky Figures $k$ with $s(k) \to 1$ or the natural language statement is true.  $s(k)$ and a natural language statement are equivalent, if and only if the resulting $K$ contains the same $k$. $s(k)$ and the natural language statement can be defined as the \textbf{Ground Truth} of a Kandinsky Pattern. 
\end{definition}

The following example shall explain the "ground truthing": 

The ground truth $gt(k) =$  "\textit{the KANDINSKY Figure has two pairs of objects with the same shape, in one pair the objects have the same colors in the other pair different colors, two pairs are always disjunctive, i.e. they do not share objects"} defines the KANDINSKYPattern $K_{gt}$,

A specific hypothesis as $h_2(k)$ = \textit{"the KANDINSKY Figure consist of two triangles with different color and two circles of  same color"} generates a KANDINSKYPattern $K_{h2}$ with $K_{gt}  \setminus K_{h2}  \neq \emptyset$, i.e. the KANDINSKYPattern of  $h_2(k)$ is missing KANDINSKY Figures which are in the KANDINSKYPattern of the ground truth. 

\subsection{Properties}

In a natural language statement about a KANDINSKYFigure humans use a series of basic concepts which are combined through logical operators, including AND, OR, NOT, etc., the following (incomplete) examples illustrate some concepts of increasing complexity.

\begin{itemize}
  \item Basic concepts given by the definition of a KANDINSKY Figure: a set of \textit{objects}, described by \textit{shape}, \textit{color},  \textit{size} and \textit{position}.
  
  \item Existence, numbers, set-relations (\textit{number}, \textit{quantity} or \textit{quantity ratios} of objects), e.g. \textit{"a KANDINSKY Figure contains four red triangles and more yellow objects than circles". }
  
  \item Spatial concepts describing the arrangement of objects, either absolute (\textit{upper}, \textit{lower}, \textit{left}, \textit{right},  \textit{between},\dots,  or relative (\textit{below}, \textit{above}, \textit{on top}, \textit{touching}, \dots), e.g. \textit{"in a KANDINSKY Figure red objects are on the left side, blue objects on the right side, and yellow objects are below blue squares".}, and concepts such as small, big, smaller as, bigger as, etc.
  
  \item Gestalt concepts e.g. \textit{closure}, \textit{symmetry}, \textit{continuity}, \textit{proximity}, \textit{similarity}, e.g. \textit{"in a KANDINSKY Figure objects are grouped in a circular manner".}
  
  \item Domain concepts, e.g. \textit{"a group of objects is perceived as a "flower"".} Remark: these are not yet found in other datasets.
\end{itemize}


\subsection{Datasets and Example Challenges}
\label{datasets}

KANDINSKYPatterns can be used as test datasets for various research questions, e.g. to address and evaluate the following topics:
\begin{enumerate}

   \item Describe classes of KANDINSKYPatterns according to their ability to be classified by machine learning algorithms in comparison to human explanation strategies. 
   \item Investigate transfer learning of concepts as \textit{numbers}, \textit{geometric positions} and \textit{Gestalt principles*)} in the classification and explanation of KANDINSKYPatterns.  
   \item Develop mapping strategies from an algorithmic classification to a known human explanation of a KANDINSKYPattern.
   \item Automatic generation of a human understandable explanation of a KANDINSKYPattern.
    
\end{enumerate}

*) Gestalt-Principles are important, because humans can recognize complex relationships even from line drawings consisting solely of contour-based shape features \cite{RezanejadEtAl:2019:Gestalt}. 

We invite the international machine learning community to experiment with our Kandinsky dataset
\url{https://github.com/human-centered-ai-lab/dat-kandinsky-patterns}, and re-use and contribute to the Kandinsky software tools \url{https://github.com/human-centered-ai-lab/app-kandinsky-pattern-generator}.

Please note that the main aim of the training datasets and the following challenges is not in the evaluation of machine learning algorithms, but most of all \textit{in explaining the successful classification by human understandable statements. }


\subsubsection{Example Challenge 1 - Objects and Shapes}
\label{challenge_nr_1}

In the challenge \textbf{Objects and Shapes} the ground truth gt(k) is defined as "\textit{in a KANDINSKY Figure small objects are arranged on big shapes same as  object shapes, in the big shape of type X, no small object of type X exists. Big square shapes only contain blue and red objects, big triangle shapes only contain yellow and red objects and big circle shape contain only yellow and blue objects"}. 

\begin{itemize}
\item \textbf{Question 1}: Which machine learning algorithm can classify Kandinsky Figures of challenge 1 - using it as benchmarks for novel NN. 
\item \textbf{Question 2:} Identify feature embeddings (colloquially "layers and regions", but layers are architectural elements and do not have a correspondence to concepts, and regions is a too general term) in the network, which corresponds to "small" and "big" shapes and the restrictions and/or limitations on object membership and color, they might be as well entangled with each other. 
\end{itemize}

Download the dataset for challenge 1 here: \url{https://tinyurl.com/KANDINSKY-C1}, \url{https://github.com/human-centered-ai-lab/dat-kandinsky-patterns/tree/master/challenge-nr-1}

\subsection{Challenge 2 - Nine Circles}
\label{challenge_nr_2}

In the challenge \textbf{Nine Circles} the set of KANDINSKY Figures consist of 9 circles arranged in a regular grid. 

The challenge is to find KANDINSKYFigures which are "almost true", i.e. they fulfill a hypothesis similar to ground truth, but are counter factual (we mean the concept - not the image), this is very important and a novel aspect of the KANDINSKYPatterns. 

\begin{itemize}
\item \textbf{Question 1}: Explain the KANDINSKYPattern in an algorithmic way, i.e. train a network which classifies KANDINSKY Figures according to ground truth of challenge 2.
\item \textbf{Question 2:} Explain the KANDINSKYPattern in natural language.
\end{itemize}

Download the dataset for challenge 2 here: \url{https://tinyurl.com/Kandinsky-C2}, \url{https://github.com/human-centered-ai-lab/dat-kandinsky-patterns/tree/master/challenge-nr-2}

\vspace{1cm}
\subsection{Challenge 3 - Blue and Yellow Circles}
\label{challenge_nr_3}

In the challenge \textbf{ Blue and Yellow Circles} the set  of all possible KANDINSKY Figures consist of equal size  blue and yellow circles. In the example figures shown in \cite{MuellerHolzinger:2019:arXivKANDINSKY}, Figure 11
shows KANDINSKY Figures according to ground truth, Figure 12 shows KANDINSKY Figures with approximately the same number of objects not belonging to the KANDINSKYPattern and Figure 13 shows KANDINSKY Figures which are "almost true", i.e. they fulfill a hypothesis similar to the ground truth.

\begin{itemize}
\item \textbf{Question 1}: Explain the KANDINSKYPattern in an algorithmic way, i.e. train a network which classifies KANDINSKY Figures according to ground truth of challenge 3.
\item \textbf{Question 2:} Explain the KANDINSKYPattern in natural language.
\end{itemize}

Download the dataset for challenge 3 here: \url{https://tinyurl.com/KANDINSKY-C3},
\url{https://github.com/human-centered-ai-lab/dat-KANDINSKY-patterns/tree/master/challenge-nr-3}

\section{Conclusion and Future Outlook}

In the following, we summarize the main features of the KANDINSKYPatterns (KP), contrast them with CLEVR, CLEVERER, CLOSURE, CURI, Bongard-Logo and V-PROM and suggest some future work. 

\begin{itemize}

\item CLEVR vs. KP:\\
KP have concepts including counting, positional, colour, shape which are similar to the basic concepts of CLEVR. Nevertheless, CLEVR encourages sequential processing of concepts, whereas higher-order concepts of KPs are hierarchical. One representative task involves arithmetic relations between different objects in a scene, for example, the number of blue and red objects equals the number of yellow objects; solving this task bases on learning to count objects.

\item CLEVERER vs KP:\\
CLEVERER is a completely different dataset than KP because it is designed for video and considers the temporal development of concepts. We work in a static environment; we don't have the temporal aspect currently. With KPs, we are currently not testing whether causality principles of a phenomenon are discovered by a neural network but this could be addressed by exploring counterfactuals.

\item CLOSURE vs KP:\\
CLOSURE is an extension of CLEVR which requires better generalization and less bias. The difference with CLEVR itself is the language principles. Textual description KP concepts and challenges as well as generalization considerations must be addressed in KP. 

\item CURI vs KP:\\
The novel component in CURI is its probabilistic grammar. This means that an image can belong to several different concepts with a certain probability. KP basic and high-order concepts also exhibit this property, therefore the construction of an appropriate grammar expressing ambiguity is a requirement.

\item Bongard-LOGO vs KP:\\
The Bongard-LOGO dataset has the most similarity with the KP-challenges. It is the only dataset where textual output is not necessary and classification without description is a big challenge. The third challenge of KP, which is thought of as the most challenging, has a corresponding task in Bongard-LOGO.  


\item V-PROM vs KP:\\ 
V-PROM extends Bongard-Logo for real images. The tasks themselves have a similar structure to Bongard-LOGO; that provides inspiration for extension of KP to real medical images. 

\end{itemize}


KANDINSKYPatterns have great potential to serve the needs of the international research community in pattern analysis and machine intelligence in general and on helping to understand concepts, ultimately from the medical domain. As other datasets that evolved over the years and variations of them are invented, KANDINSKYPatterns will need to adapt evolutionary as well: The first enhancement is in the definition of a grammar and/or a domain-specific language that encompasses the basic concepts that are relevant to a medical diagnosis (see also \cite{DickinsonPizlo:2013:Shape}). The dependencies between them, as well as their hierarchical compositional structure, needs to be clearly defined. Aspects including ambiguity and data that could be produced by several concepts, have to be supported by the expressive capabilities of the designed language. At the same time, this textual representation must correspond to the way people in general and medical professionals, in particular, express their understanding. 

All technical solutions presented in section \ref{TechnicalSolutions} can be investigated utilizing the KP dataset. Their performance needs to be carefully documented before developing new AI solutions tailored to KP needs. New neural network architectures or even hybrid methods that incorporate symbolic knowledge need to be invented. In this regard, various xAI methods will be very helpful in shedding light on the reasons behind the successes and shortcomings of the tested methods. 

Dedicated splits that systematically test the generalization and compositional abilities of state-of-the-art neural networks on the KANDINSKYPatterns dataset, as well as considerations about the distributions presented in the training and test phase (input images and text), need to be implemented. The repository allows developers and data scientists unconstrained generation of verified data,  but few-shot learning is one of the important goals of recent neural network solutions. This leads naturally to a comparison with human-level performance as far as correct concept recognition, expression in natural language and human generalization and disentanglement abilities. The quantification of how well humans describe and discriminate KP as well as the comparison to the ground truth is an important metric that will be used for comparison to the AI solutions performance (without human performance being necessarily better, as seen in \cite{Teney:2020:VPROM}).

\section*{Acknowledgements}

This work has received funding by the Austrian Science Fund (FWF), Project: P-32554 ``A reference model for explainable Artificial Intelligence in the medical domain''.

\ifCLASSOPTIONcaptionsoff
\fi



%

{
\bibliographystyle{IEEEtran}
\bibliography{references}

\begin{thebibliography}{10}
\providecommand{\url}[1]{#1}
\csname url@samestyle\endcsname
\providecommand{\newblock}{\relax}
\providecommand{\bibinfo}[2]{#2}
\providecommand{\BIBentrySTDinterwordspacing}{\spaceskip=0pt\relax}
\providecommand{\BIBentryALTinterwordstretchfactor}{4}
\providecommand{\BIBentryALTinterwordspacing}{\spaceskip=\fontdimen2\font plus
\BIBentryALTinterwordstretchfactor\fontdimen3\font minus
  \fontdimen4\font\relax}
\providecommand{\BIBforeignlanguage}[2]{{%
\expandafter\ifx\csname l@#1\endcsname\relax
\typeout{** WARNING: IEEEtran.bst: No hyphenation pattern has been}%
\typeout{** loaded for the language `#1'. Using the pattern for}%
\typeout{** the default language instead.}%
\else
\language=\csname l@#1\endcsname
\fi
#2}}
\providecommand{\BIBdecl}{\relax}
\BIBdecl

\bibitem{Jain:2000:statistical}
A.~K. Jain, R.~P.~W. Duin, and J.~Mao, ``Statistical pattern recognition: A
  review,'' \emph{Pattern Analysis and Machine Intelligence, IEEE Transactions
  on}, vol.~22, no.~1, pp. 4--37, 2000.

\bibitem{EstevaThrun:2017:DermaNN}
A.~Esteva, B.~Kuprel, R.~A. Novoa, J.~Ko, S.~M. Swetter, H.~M. Blau, and
  S.~Thrun, ``Dermatologist-level classification of skin cancer with deep
  neural networks,'' \emph{Nature}, vol. 542, no. 7639, pp. 115--118, 2017.

\bibitem{ArdilaEtAl:2019:DeepSuccess}
D.~Ardila, A.~P. Kiraly, S.~Bharadwaj, B.~Choi, J.~J. Reicher, L.~Peng, D.~Tse,
  M.~Etemadi, W.~Ye, and G.~Corrado, ``End-to-end lung cancer screening with
  three-dimensional deep learning on low-dose chest computed tomography,''
  \emph{Nature medicine}, vol.~25, no.~6, pp. 954--961, 2019.

\bibitem{Esteva:2021:DeepVisionNature}
A.~Esteva, K.~Chou, S.~Yeung, N.~Naik, A.~Madani, A.~Mottaghi, Y.~Liu,
  E.~Topol, J.~Dean, and R.~Socher, ``Deep learning-enabled medical computer
  vision,'' \emph{Nature Digital Medicine}, vol.~4, no.~1, pp. 1--9, 2021.

\bibitem{MisraVanDerMaaten:2018:LearningAsking}
I.~Misra, R.~Girshick, R.~Fergus, M.~Hebert, A.~Gupta, and L.~Van Der~Maaten,
  ``Learning by asking questions,'' in \emph{Proceedings of the IEEE Conference
  on Computer Vision and Pattern Recognition}.\hskip 1em plus 0.5em minus
  0.4em\relax IEEE, 2018.

\bibitem{SammutBanerji:1986:FirstOrderLogic}
C.~Sammut and R.~B. Banerji, ``Learning concepts by asking questions,'' in
  \emph{Machine learning: An artificial intelligence approach, Volume II},
  R.~S. Michalski, J.~G. Carbonell, and T.~M. Mitchell, Eds.\hskip 1em plus
  0.5em minus 0.4em\relax Los Altos (CA): Morgan Kaufmann, 1986, pp. 167--192.

\bibitem{Tenenbaum:2020:Human-Level}
K.~Ota, D.~K. Jha, D.~Romeres, J.~van Baar, K.~A. Smith, T.~Semitsu, T.~Oiki,
  A.~Sullivan, D.~Nikovski, and J.~B. Tenenbaum, ``Towards human-level learning
  of complex physical puzzles,'' \emph{arXiv:2011.07193}, 2020.

\bibitem{Barsalou:1983:AdHocCategories}
L.~W. Barsalou, ``Ad hoc categories,'' \emph{Memory and Cognition}, vol.~11,
  no.~3, pp. 211--227, 1983.

\bibitem{Piantadosi2016-bx}
S.~T. Piantadosi, J.~B. Tenenbaum, and N.~D. Goodman,
  ``\BIBforeignlanguage{en}{The logical primitives of thought: Empirical
  foundations for compositional cognitive models},''
  \emph{\BIBforeignlanguage{en}{Psychological Review}}, vol. 123, no.~4, pp.
  392--424, Jul. 2016.

\bibitem{Kandinsky:1926:PunktLinie}
W.~Kandinsky, \emph{Punkt und Linie zu Fl\"{a}che: Beitrag zur Analyse der
  malerischen Elemente.}\hskip 1em plus 0.5em minus 0.4em\relax Muenchen:
  Albert Langen, 1926.

\bibitem{HubelWiesel:1959:Cat}
D.~H. Hubel and T.~N. Wiesel, ``Receptive fields of single neurons in the cat's
  striate cortex,'' \emph{The Journal of physiology}, vol. 148, no.~3, pp.
  574--591, 1959.

\bibitem{LeCunBengioHinton:2015:DeepLearningNature}
Y.~LeCun, Y.~Bengio, and G.~Hinton, ``Deep learning,'' \emph{Nature}, vol. 521,
  no. 7553, pp. 436--444, 2015.

\bibitem{Antol:2015:VQAVersion_1}
S.~Antol, A.~Agrawal, J.~Lu, M.~Mitchell, D.~Batra, C.~L. Zitnick, and
  D.~Parikh, ``Vqa: Visual question answering,'' in \emph{Proceedings of the
  IEEE international conference on computer vision}, 2015, pp. 2425--2433.

\bibitem{Makino:2020:DifferencesHumanMachineMedical}
T.~Makino, S.~Jastrzebski, W.~Oleszkiewicz, C.~Chacko, R.~Ehrenpreis,
  N.~Samreen, C.~Chhor, E.~Kim, J.~Lee, K.~Pysarenko \emph{et~al.},
  ``Differences between human and machine perception in medical diagnosis,''
  \emph{arXiv:2011.14036}, 2020.

\bibitem{Koller:2009:ProbabilisticGraphicalModels}
D.~Koller and N.~Friedman, \emph{Probabilistic graphical models: principles and
  techniques}.\hskip 1em plus 0.5em minus 0.4em\relax MIT press, 2009.

\bibitem{Saranti:2019:LearningCompetencePGM}
A.~Saranti, B.~Taraghi, M.~Ebner, and A.~Holzinger, ``Insights into learning
  competence through probabilistic graphical models,'' in \emph{International
  cross-domain conference for machine learning and knowledge extraction}.\hskip
  1em plus 0.5em minus 0.4em\relax Springer, 2019, pp. 250--271.

\bibitem{Lin:2014:MicrosoftCOCO}
T.-Y. Lin, M.~Maire, S.~Belongie, J.~Hays, P.~Perona, D.~Ramanan,
  P.~Doll{\'a}r, and C.~L. Zitnick, ``Microsoft coco: Common objects in
  context,'' in \emph{European conference on computer vision}.\hskip 1em plus
  0.5em minus 0.4em\relax Springer, 2014, pp. 740--755.

\bibitem{Chen:2015:MicrosoftCOCOCaptions}
X.~Chen, H.~Fang, T.-Y. Lin, R.~Vedantam, S.~Gupta, P.~Doll{\'a}r, and C.~L.
  Zitnick, ``Microsoft coco captions: Data collection and evaluation server,''
  \emph{arXiv:1504.00325}, 2015.

\bibitem{Kulkarni:2013:Babytalk}
G.~Kulkarni, V.~Premraj, V.~Ordonez, S.~Dhar, S.~Li, Y.~Choi, A.~C. Berg, and
  T.~L. Berg, ``Babytalk: Understanding and generating simple image
  descriptions,'' \emph{IEEE Transactions on Pattern Analysis and Machine
  Intelligence}, vol.~35, no.~12, pp. 2891--2903, 2013.

\bibitem{Karpathy:2015:DeepVisualSemanticAlignments}
A.~Karpathy and L.~Fei-Fei, ``Deep visual-semantic alignments for generating
  image descriptions,'' in \emph{Proceedings of the IEEE conference on computer
  vision and pattern recognition}, 2015, pp. 3128--3137.

\bibitem{Hendricks:2016:GeneratingVisualExplanations}
L.~A. Hendricks, Z.~Akata, M.~Rohrbach, J.~Donahue, B.~Schiele, and T.~Darrell,
  ``Generating visual explanations,'' in \emph{European Conference on Computer
  Vision}.\hskip 1em plus 0.5em minus 0.4em\relax Springer, 2016, pp. 3--19.

\bibitem{Lai:2019:Contextual}
F.~Lai, N.~Xie, D.~Doran, and A.~Kadav, ``Contextual grounding of natural
  language entities in images,'' \emph{arXiv:1911.02133}, 2019.

\bibitem{Lapuschkin:2019:CleverHans}
S.~Lapuschkin, S.~W{\"a}ldchen, A.~Binder, G.~Montavon, W.~Samek, and K.-R.
  M{\"u}ller, ``Unmasking clever hans predictors and assessing what machines
  really learn,'' \emph{Nature communications}, vol.~10, no.~1, pp. 1--8, 2019.

\bibitem{Krishna:2017:VisualGenome}
R.~Krishna, Y.~Zhu, O.~Groth, J.~Johnson, K.~Hata, J.~Kravitz, S.~Chen,
  Y.~Kalantidis, L.-J. Li, D.~A. Shamma \emph{et~al.}, ``Visual genome:
  Connecting language and vision using crowdsourced dense image annotations,''
  \emph{International journal of computer vision}, vol. 123, no.~1, pp. 32--73,
  2017.

\bibitem{Mao:2019:NeuroSymbolicConceptLearner}
J.~Mao, C.~Gan, P.~Kohli, J.~B. Tenenbaum, and J.~Wu, ``The neuro-symbolic
  concept learner: Interpreting scenes, words, and sentences from natural
  supervision,'' \emph{arXiv:1904.12584}, 2019.

\bibitem{Agrawal:2017:CVQAACompositionalSplit}
A.~Agrawal, A.~Kembhavi, D.~Batra, and D.~Parikh, ``C-vqa: A compositional
  split of the visual question answering (vqa) v1.0 dataset,''
  \emph{arXiv:1704.08243}, 2017.

\bibitem{Yang:2020:JustAskVideoQA}
A.~Yang, A.~Miech, J.~Sivic, I.~Laptev, and C.~Schmid, ``Just ask: Learning to
  answer questions from millions of narrated videos,'' \emph{arXiv:2012.00451},
  2020.

\bibitem{Kojima:2020:WhatisLearnedVQA}
N.~Kojima, H.~Averbuch-Elor, A.~M. Rush, and Y.~Artzi, ``What is learned in
  visually grounded neural syntax acquisition,'' \emph{arXiv:2005.01678}, 2020.

\bibitem{Torralba:2011:DatasetBias}
A.~Torralba, A.~A. Efros \emph{et~al.}, ``Unbiased look at dataset bias.'' in
  \emph{CVPR}, vol.~1, no.~2.\hskip 1em plus 0.5em minus 0.4em\relax Citeseer,
  2011, p.~7.

\bibitem{Bahdanau:2019:SystematicGeneralizationWhatIsRequired}
D.~Bahdanau, S.~Murty, M.~Noukhovitch, T.~Nguyen, H.~D. Vries, and A.~C.
  Courville, ``Systematic generalization: What is required and can it be
  learned?'' \emph{arXiv:1811.12889}, 2019.

\bibitem{Andreas:2016:NeuralModuleNetworks}
J.~Andreas, M.~Rohrbach, T.~Darrell, and D.~Klein, ``Neural module networks,''
  in \emph{Proceedings of the IEEE conference on computer vision and pattern
  recognition}, 2016, pp. 39--48.

\bibitem{HudsonManning:2018:MACCompositionalAttention}
D.~A. Hudson and C.~D. Manning, ``Compositional attention networks for machine
  reasoning,'' \emph{arXiv:1803.03067}, 2018.

\bibitem{Santoro:2017:SimpleNNRelationalReasoning}
A.~Santoro, D.~Raposo, D.~G. Barrett, M.~Malinowski, R.~Pascanu, P.~Battaglia,
  and T.~Lillicrap, ``A simple neural network module for relational
  reasoning,'' in \emph{Advances in neural information processing systems},
  2017, pp. 4967--4976.

\bibitem{Perez:2018:FiLM}
E.~Perez, F.~Strub, H.~De~Vries, V.~Dumoulin, and A.~Courville, ``Film: Visual
  reasoning with a general conditioning layer,'' in \emph{Proceedings of the
  AAAI Conference on Artificial Intelligence}, vol.~32, no.~1, 2018.

\bibitem{Keysers:2020:MeasuringCompositionalGeneralization}
D.~Keysers, N.~Sch{\"a}rli, N.~Scales, H.~Buisman, D.~Furrer, S.~Kashubin,
  N.~Momchev, D.~Sinopalnikov, L.~Stafiniak, T.~Tihon, D.~Tsarkov, X.~Wang,
  M.~van Zee, and O.~Bousquet, ``Measuring compositional generalization: A
  comprehensive method on realistic data,'' \emph{arXiv:1912.09713}, 2020.

\bibitem{Chung:1989:DistanceProbabilityDistributions}
J.~Chung, P.~Kannappan, C.~Ng, and P.~Sahoo, ``Measures of distance between
  probability distributions,'' \emph{Journal of mathematical analysis and
  applications}, vol. 138, no.~1, pp. 280--292, 1989.

\bibitem{Johnson:2017:Clevr}
J.~Johnson, B.~Hariharan, L.~van~der Maaten, L.~Fei-Fei, C.~Lawrence~Zitnick,
  and R.~Girshick, ``Clevr: A diagnostic dataset for compositional language and
  elementary visual reasoning,'' in \emph{Proceedings of the IEEE Conference on
  Computer Vision and Pattern Recognition}, 2017, pp. 2901--2910.

\bibitem{Kim:2018:NotSoCLEVRL}
J.~Kim, M.~Ricci, and T.~Serre, ``Not-so-clevr: learning same–different
  relations strains feedforward neural networks,'' \emph{Interface Focus},
  vol.~8, 2018.

\bibitem{Bahdanau:2019:Closure}
D.~Bahdanau, H.~de~Vries, T.~J. O'Donnell, S.~Murty, P.~Beaudoin, Y.~Bengio,
  and A.~Courville, ``Closure: Assessing systematic generalization of clevr
  models,'' \emph{arXiv:1912.05783}, 2019.

\bibitem{Johnson:2017:ClevrHumans}
J.~Johnson, B.~Hariharan, L.~Van Der~Maaten, J.~Hoffman, L.~Fei-Fei,
  C.~Lawrence~Zitnick, and R.~Girshick, ``Inferring and executing programs for
  visual reasoning,'' in \emph{Proceedings of the IEEE International Conference
  on Computer Vision}, 2017, pp. 2989--2998.

\bibitem{Yi:2019:Clevrer}
K.~Yi, C.~Gan, Y.~Li, P.~Kohli, J.~Wu, A.~Torralba, and J.~B. Tenenbaum,
  ``Clevrer: Collision events for video representation and reasoning,''
  \emph{arXiv:1910.01442}, 2019.

\bibitem{TverskyKahnemann:1974:JudgementUncertainty}
A.~Tversky and D.~Kahneman, ``Judgment under uncertainty: Heuristics and
  biases,'' \emph{Science}, vol. 185, no. 4157, pp. 1124--1131, 1974.

\bibitem{Kahneman:2001:thinkingFastSlow}
D.~Kahneman, \emph{Thinking, fast and slow}.\hskip 1em plus 0.5em minus
  0.4em\relax New York: Macmillan, 2011.

\bibitem{Vedantam:2020:CuriConceptLearningUncertainty}
R.~Vedantam, A.~Szlam, M.~Nickel, A.~Morcos, and B.~Lake, ``Curi: A benchmark
  for productive concept learning under uncertainty,'' \emph{arXiv:2010.02855},
  2020.

\bibitem{Kipf:2019:ContrastiveLearning}
T.~Kipf, E.~van~der Pol, and M.~Welling, ``Contrastive learning of structured
  world models,'' \emph{arXiv:1911.12247}, 2019.

\bibitem{Carpenter:1990:RavenProgressiveMatricesTest}
P.~A. Carpenter, M.~A. Just, and P.~Shell, ``What one intelligence test
  measures: a theoretical account of the processing in the raven progressive
  matrices test.'' \emph{Psychological review}, vol.~97, no.~3, p. 404, 1990.

\bibitem{Raven:2000:Rpms}
J.~Raven, ``The raven's progressive matrices: change and stability over culture
  and time,'' \emph{Cognitive psychology}, vol.~41, no.~1, pp. 1--48, 2000.

\bibitem{Zhang:2019:RavenDataset}
C.~Zhang, F.~Gao, B.~Jia, Y.~Zhu, and S.-C. Zhu, ``Raven: A dataset for
  relational and analogical visual reasoning,'' in \emph{Proceedings of the
  IEEE/CVF Conference on Computer Vision and Pattern Recognition}, 2019, pp.
  5317--5327.

\bibitem{Teney:2020:VPROM}
D.~Teney, P.~Wang, J.~Cao, L.~Liu, C.~Shen, and A.~van~den Hengel, ``V-prom: A
  benchmark for visual reasoning using visual progressive matrices.'' in
  \emph{AAAI}, 2020, pp. 12\,071--12\,078.

\bibitem{Nie:2020:BongardLogo}
W.~Nie, Z.~Yu, L.~Mao, A.~B. Patel, Y.~Zhu, and A.~Anandkumar, ``Bongard-logo:
  A new benchmark for human-level concept learning and reasoning,''
  \emph{Advances in Neural Information Processing Systems}, vol.~33, 2020.

\bibitem{Bongard:1968:TheRecognitionProblem}
M.~M. Bongard, ``The recognition problem,'' FOREIGN TECHNOLOGY DIV
  WRIGHT-PATTERSON AFB OHIO, Tech. Rep., 1968.

\bibitem{Bongard:1970:PatternRecognition}
\BIBentryALTinterwordspacing
M.~Bongard, J.~Hawkins, and T.~Cheron, \emph{Pattern Recognition}, ser.
  Problema uznavaniia.\hskip 1em plus 0.5em minus 0.4em\relax Spartan Books,
  1970. [Online]. Available:
  \url{https://books.google.at/books?id=vY1QAAAAMAAJ}
\BIBentrySTDinterwordspacing

\bibitem{Harvey:1997:Logo}
B.~Harvey, \emph{Computer Science Logo Style: Symbolic Computing}.\hskip 1em
  plus 0.5em minus 0.4em\relax MIT press, 1997, vol.~1.

\bibitem{Kuhnle:2017:Shapeworld}
A.~Kuhnle and A.~Copestake, ``Shapeworld-a new test methodology for multimodal
  language understanding,'' \emph{arXiv:1704.04517}, 2017.

\bibitem{Weston:2015:bAbi}
J.~Weston, A.~Bordes, S.~Chopra, A.~M. Rush, B.~van Merri{\"e}nboer, A.~Joulin,
  and T.~Mikolov, ``Towards ai-complete question answering: A set of
  prerequisite toy tasks,'' \emph{arXiv:1502.05698}, 2015.

\bibitem{Bakhtin2019-sy}
A.~Bakhtin, L.~van~der Maaten, J.~Johnson, L.~Gustafson, and R.~Girshick,
  ``{PHYRE}: A new benchmark for physical reasoning,'' \emph{arXiv}, Aug. 2019.

\bibitem{Saxton:2019:MathematicalReasoning}
D.~Saxton, E.~Grefenstette, F.~Hill, and P.~Kohli, ``Analysing mathematical
  reasoning abilities of neural models,'' \emph{arXiv:1904.01557}, 2019.

\bibitem{Andreas:2016:ComposeNeuralNetworksQA}
J.~Andreas, M.~Rohrbach, T.~Darrell, and D.~Klein, ``Learning to compose neural
  networks for question answering,'' \emph{arXiv:1601.01705}, 2016.

\bibitem{Hu:2017:LearningToReason}
R.~Hu, J.~Andreas, M.~Rohrbach, T.~Darrell, and K.~Saenko, ``Learning to
  reason: End-to-end module networks for visual question answering,'' in
  \emph{Proceedings of the IEEE International Conference on Computer Vision},
  2017, pp. 804--813.

\bibitem{Kim:2018:ProgressieModuleNetworks}
S.~W. Kim, M.~Tapaswi, and S.~Fidler, ``Visual reasoning by progressive module
  networks,'' \emph{arXiv:1806.02453}, 2018.

\bibitem{Sutton:2018:ReinforcementLearning}
R.~S. Sutton and A.~G. Barto, \emph{Reinforcement learning: An
  introduction}.\hskip 1em plus 0.5em minus 0.4em\relax MIT press, 2018.

\bibitem{Yi:2018:NeuralSymbolicVQA}
K.~Yi, J.~Wu, C.~Gan, A.~Torralba, P.~Kohli, and J.~Tenenbaum,
  ``Neural-symbolic vqa: Disentangling reasoning from vision and language
  understanding,'' in \emph{Advances in Neural Information Processing Systems},
  2018, pp. 1031--1042.

\bibitem{Williams:1992:REINFORCE}
R.~J. Williams, ``Simple statistical gradient-following algorithms for
  connectionist reinforcement learning,'' \emph{Machine learning}, vol.~8, no.
  3-4, pp. 229--256, 1992.

\bibitem{Han:2019:VisualConceptMetaconcept}
C.~Han, J.~Mao, C.~Gan, J.~Tenenbaum, and J.~Wu, ``Visual concept-metaconcept
  learning,'' in \emph{Advances in Neural Information Processing Systems},
  2019, pp. 5002--5013.

\bibitem{Zhao:2020:VisuallyGroundedCompoundPCFGs}
Y.~Zhao and I.~Titov, ``Visually grounded compound pcfgs,''
  \emph{arXiv:2009.12404}, 2020.

\bibitem{Shi:2019:VisuallyGroundedNeuralSyntaxAcquisition}
H.~Shi, J.~Mao, K.~Gimpel, and K.~Livescu, ``Visually grounded neural syntax
  acquisition,'' \emph{arXiv:1906.02890}, 2019.

\bibitem{Barrett2018-xv}
D.~G.~T. Barrett, F.~Hill, A.~Santoro, A.~S. Morcos, and T.~Lillicrap,
  ``Measuring abstract reasoning in neural networks,'' \emph{arXiv}, Jul. 2018.

\bibitem{Saqur:2020:MultimodalGNNs}
R.~Saqur and K.~Narasimhan, ``Multimodal graph networks for compositional
  generalization in visual question answering,'' \emph{Advances in Neural
  Information Processing Systems}, vol.~33, 2020.

\bibitem{Zhou:2019:CgcNet}
Y.~Zhou, S.~Graham, N.~Alemi~Koohbanani, M.~Shaban, P.-A. Heng, and N.~Rajpoot,
  ``Cgc-net: Cell graph convolutional network for grading of colorectal cancer
  histology images,'' in \emph{Proceedings of the IEEE International Conference
  on Computer Vision Workshops}, 2019, pp. 0--0.

\bibitem{Schnake:2020:xaiGNNs}
T.~Schnake, O.~Eberle, J.~Lederer, S.~Nakajima, K.~T. Sch{\"u}tt, K.-R.
  M{\"u}ller, and G.~Montavon, ``Xai for graphs: Explaining graph neural
  network predictions by identifying relevant walks,'' \emph{arXiv:2006.03589},
  2020.

\bibitem{Muzio:2020:BioNetworkAnalysisDL}
G.~Muzio, L.~O’Bray, and K.~Borgwardt, ``Biological network analysis with
  deep learning,'' \emph{Briefings in Bioinformatics}, 2020.

\bibitem{Kori:2020:DNNstoConceptGraphs}
A.~Kori, P.~Natekar, G.~Krishnamurthi, and B.~Srinivasan, ``Abstracting deep
  neural networks into concept graphs for concept level interpretability,''
  \emph{arXiv:2008.06457}, 2020.

\bibitem{SarantiEtAl:2019:graphicalModels}
A.~Saranti, B.~Taraghi, M.~Ebner, and A.~Holzinger, ``Insights into learning
  competence through probabilistic graphical models,'' in \emph{Machine
  Learning and Knowledge Extraction, Lecture Notes in Computer Science},
  A.~Holzinger, P.~Kieseberg, A.~M. Tjoa, and E.~Weippl, Eds.\hskip 1em plus
  0.5em minus 0.4em\relax Cham: Springer/Nature, 2019, pp. 250--271.

\bibitem{Holzinger:2016:InteractiveMLHumanInTheLoop}
A.~Holzinger, ``Interactive machine learning for health informatics: when do we
  need the human-in-the-loop?'' \emph{Brain Informatics}, vol.~3, no.~2, pp.
  119--131, 2016.

\bibitem{Sato:2015:EulerDiagrams}
Y.~Sato, S.~Masuda, Y.~Someya, T.~Tsujii, and S.~Watanabe, ``An fmri analysis
  of the efficacy of euler diagrams in logical reasoning,'' in \emph{2015 IEEE
  Symposium on Visual Languages and Human-Centric Computing (VL/HCC)}.\hskip
  1em plus 0.5em minus 0.4em\relax IEEE, 2015, pp. 143--151.

\bibitem{Wang:2020:AbstractDiagrammaticReasoningMutliplexGN}
D.~Wang, M.~Jamnik, and P.~Lio, ``Abstract diagrammatic reasoning with
  multiplex graph networks,'' \emph{arXiv:2006.11197}, 2020.

\bibitem{Yang:2020:ObjCentricVisualReasoning}
J.~Yang, J.~Mao, J.~Wu, D.~Parikh, D.~D. Cox, J.~B. Tenenbaum, and C.~Gan,
  ``Object-centric diagnosis of visual reasoning,'' \emph{arXiv:2012.11587},
  2020.

\bibitem{Lamb:2020:GNNsSymbolicAI}
L.~Lamb, A.~Garcez, M.~Gori, M.~Prates, P.~Avelar, and M.~Vardi, ``Graph neural
  networks meet neural-symbolic computing: A survey and perspective,''
  \emph{arXiv:2003.00330}, 2020.

\bibitem{Holzinger:2021:MultiModalCausabilityGNN}
A.~Holzinger, B.~Malle, A.~Saranti, and B.~Pfeifer, ``Towards multi-modal
  causability with graph neural networks enabling information fusion for
  explainable ai,'' \emph{Information Fusion}, 2021.

\bibitem{Chen:2020:CounterfactualVQA}
L.~Chen, X.~Yan, J.~Xiao, H.~Zhang, S.~Pu, and Y.~Zhuang, ``Counterfactual
  samples synthesizing for robust visual question answering,'' in
  \emph{Proceedings of the IEEE/CVF Conference on Computer Vision and Pattern
  Recognition}, 2020, pp. 10\,800--10\,809.

\bibitem{Gokhale:2020:MutantVQA}
T.~Gokhale, P.~Banerjee, C.~Baral, and Y.~Yang, ``Mutant: A training paradigm
  for out-of-distribution generalization in visual question answering,''
  \emph{arXiv:2009.08566}, 2020.

\bibitem{VonKluegelgen:2020:CausalGenerativeSceneModels}
J.~von K{\"u}gelgen, I.~Ustyuzhaninov, P.~Gehler, M.~Bethge, and
  B.~Sch{\"o}lkopf, ``Towards causal generative scene models via competition of
  experts,'' \emph{arXiv:2004.12906}, 2020.

\bibitem{Burgess:2019:Monet}
C.~P. Burgess, L.~Matthey, N.~Watters, R.~Kabra, I.~Higgins, M.~Botvinick, and
  A.~Lerchner, ``Monet: Unsupervised scene decomposition and representation,''
  \emph{arXiv:1901.11390}, 2019.

\bibitem{HudsonManning:2019l:NeuralStateMachineProbabilisticGraph}
D.~Hudson and C.~D. Manning, ``Learning by abstraction: The neural state
  machine,'' in \emph{Advances in Neural Information Processing Systems}, 2019,
  pp. 5901--5914.

\bibitem{Misra:2018:LearningByAskingQuestions}
I.~Misra, R.~Girshick, R.~Fergus, M.~Hebert, A.~Gupta, and L.~Van Der~Maaten,
  ``Learning by asking questions,'' in \emph{Proceedings of the IEEE Conference
  on Computer Vision and Pattern Recognition}, 2018, pp. 11--20.

\bibitem{Rothman:2020:HandsOnXAI}
\BIBentryALTinterwordspacing
D.~Rothman, \emph{Hands-On Explainable AI (XAI) with Python: Interpret,
  visualize, explain, and integrate reliable AI for fair, secure, and
  trustworthy AI apps}.\hskip 1em plus 0.5em minus 0.4em\relax Packt
  Publishing, 2020. [Online]. Available:
  \url{https://books.google.at/books?id=2f30DwAAQBAJ}
\BIBentrySTDinterwordspacing

\bibitem{Holzinger:2019:Kandinsky}
A.~Holzinger, M.~Kickmeier-Rust, and H.~M{\"u}ller, ``Kandinsky patterns as
  iq-test for machine learning,'' in \emph{International Cross-Domain
  Conference for Machine Learning and Knowledge Extraction}.\hskip 1em plus
  0.5em minus 0.4em\relax Springer, 2019, pp. 1--14.

\bibitem{MuellerHolzinger:2019:arXivKANDINSKY}
H.~Mueller and A.~Holzinger, ``Kandinsky patterns,'' \emph{arXiv:1906.00657},
  2019.

\bibitem{RezanejadEtAl:2019:Gestalt}
M.~Rezanejad, G.~Downs, J.~Wilder, D.~B. Walther, A.~Jepson, S.~Dickinson, and
  K.~Siddiqi, ``Gestalt-based contour weights improve scene categorization by
  cnns,'' in \emph{Conference on Cognitive Computational Neuroscience (CCN
  2019)}, 2019.

\bibitem{DickinsonPizlo:2013:Shape}
S.~Dickinson and Z.~Pizlo, \emph{Shape perception in human and computer
  vision}.\hskip 1em plus 0.5em minus 0.4em\relax London: Springer, 2013.

\end{thebibliography}
}



%

\vspace*{-2cm}

\begin{IEEEbiography}[{\includegraphics[width=1in,height=1.25in,clip,keepaspectratio]{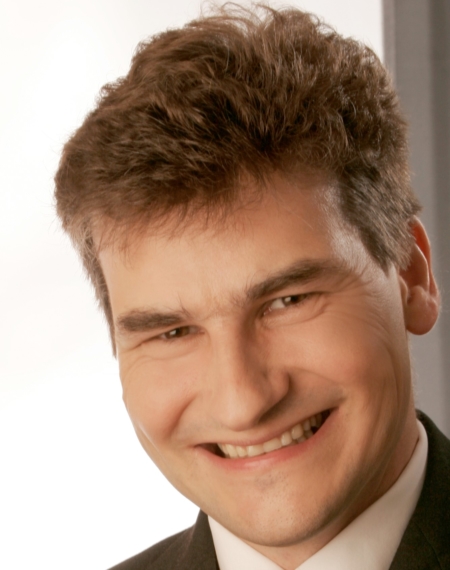}}]{Andreas Holzinger} (M'00) is Visiting Prof. for xAI at the University of Alberta, Canada since 2019 and head of the Human-Centered AI Lab at the Medical University Graz, Austria. He received his PhD in cognitive science from Graz University in 1998 and his second PhD in computer science from Graz University of Technology in 2003. He is Austrian representative in the IFIP TC 12 AI and member of WG Computational Intelligence, the ACM, IEEE, GI, AAAI, and in the board of AI made in Germany 2030.
\end{IEEEbiography}

\vspace*{-2cm}

\begin{IEEEbiography}[{\includegraphics[width=1in,height=1.25in,clip,keepaspectratio]{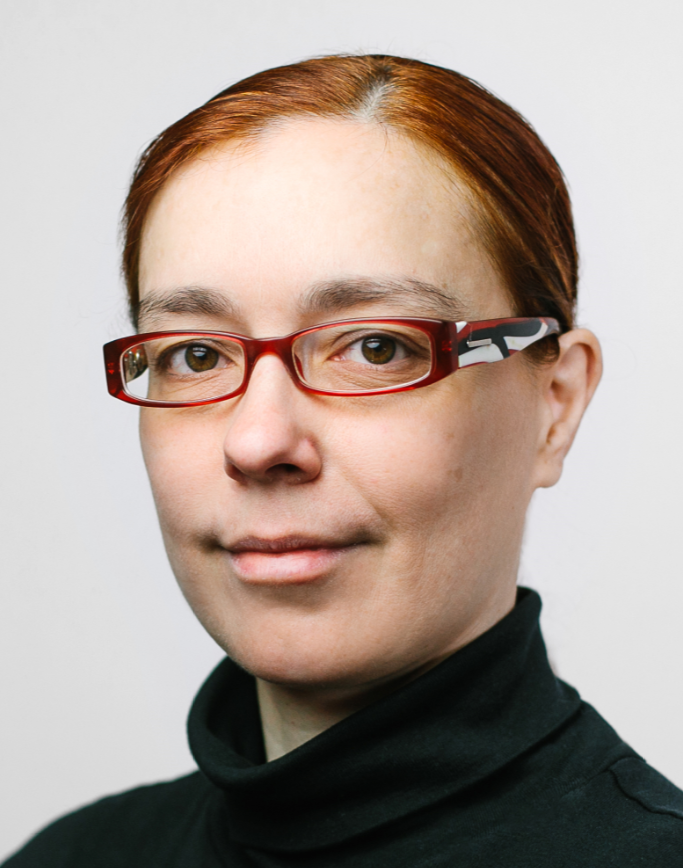}}]{Anna Saranti} is currently pursuing her PhD in the field of explainable AI. Anna received her MSc from Graz University of Technology in 2019 with a thesis on Applying Probabilistic Graphical Models and Deep Reinforcement Learning. 
\end{IEEEbiography}

\vspace*{-2cm}

\begin{IEEEbiography}[{\includegraphics[width=1in,height=1.25in,clip,keepaspectratio]{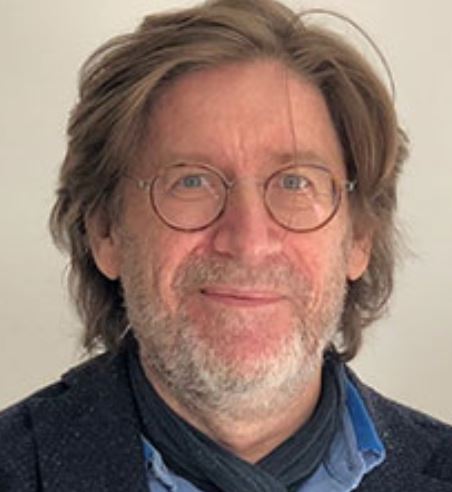}}]{Heimo Mueller} is head of the Information Science and Machine Learning group at the Diagnostic and Research Institute for Pathology of the Medical University Graz. Heimo received his PhD in 1995 with a thesis on data space semantics from Vienna University of Technology. 
\end{IEEEbiography}

\end{document}